\def\year{2022}\relax
\documentclass[letterpaper]{article} 
\usepackage{aaai22}  
\usepackage{times}  
\usepackage{helvet}  
\usepackage{courier}  
\usepackage[hyphens]{url}  
\usepackage{graphicx} 
\usepackage{natbib}
\usepackage{caption}
\DeclareCaptionStyle{ruled}{labelfont=normalfont,labelsep=colon,strut=off} 
\usepackage{algorithm}
\usepackage{algorithmic}

\usepackage{newfloat}
\usepackage{listings}
\floatstyle{ruled}
\newfloat{listing}{tb}{lst}{}
\floatname{listing}{Listing}
\usepackage{cite}
\usepackage{booktabs}               
\usepackage{subcaption}
\usepackage{xcolor}
\usepackage{wrapfig}

\usepackage{amsmath}

\usepackage{bm}

\usepackage{amssymb}
\usepackage[T1]{fontenc}

\newcommand{\Blue}        {\textcolor{blue}}

\newcommand{\field}[1]{\mathbb{#1}}
\newcommand{\R}{\field{R}}                              

\newcommand{\mat}[1]{\boldsymbol{#1}}

\newcommand{\bobs}      {\mat{o}}
\newcommand{\bcamera}   {\mat{c}}
\newcommand{\bmea}      {\mat{m}}
\newcommand{\bway}      {\mat{w}}
\newcommand{\broute}    {\mat{g}}

\newcommand{\by}        {\mat{y}}
\newcommand{\ba}        {\mat{a}}
\newcommand{\bb}        {\mat{b}}
\newcommand{\bk}        {\mat{k}}
\newcommand{\bq}        {\mat{q}}
\newcommand{\bv}        {\mat{v}}
\newcommand{\bx}        {\mat{x}}
\newcommand{\bz}        {\mat{z}}

\newcommand{\bW}        {\mat{W}}

\newcommand{\mL}        {\mathcal{L}}

\newcommand{\rrou}      {\mathrm{rou}}
\newcommand{\rcam}      {\mathrm{cam}}
\newcommand{\rlig}      {\mathrm{lig}}
\newcommand{\rste}      {\mathrm{ste}}
\newcommand{\rthro}     {\mathrm{thro}}

\newcommand{\rfull}     {\mathrm{full}}
\newcommand{\rclip}     {\mathrm{clip}}
\newcommand{\rst}     {\mathrm{st}}

\newcommand{\eat}[1]{}                                 

\title{CADRE: A Cascade Deep Reinforcement Learning Framework for \\ Vision-based Autonomous Urban Driving}
\author{
    Yinuo Zhao\equalcontrib\textsuperscript{\rm 1},
    Kun Wu\equalcontrib\textsuperscript{\rm 2},
    Zhiyuan Xu\textsuperscript{\rm 3},
    Zhengping Che\textsuperscript{\rm 3},\\
    Qi Lu\textsuperscript{\rm 3},
    Jian Tang\thanks{Corresponding author.}\textsuperscript{\rm 3},
    Chi Harold Liu\textsuperscript{\rm 1}
}
\affiliations {
    \textsuperscript{\rm 1} Beijing Institute of Technology \ 
    \textsuperscript{\rm 2} Syracuse University \ 
    \textsuperscript{\rm 3} Midea Group \\
    \textsuperscript{\rm 1} 3120191078@bit.edu.cn \ \textsuperscript{\rm 2} kwu102@syr.edu \ \textsuperscript{\rm 3} \{xuzy70,chezp,luqi8,tangjian22\}@midea.com \\ \textsuperscript{\rm 1} liuchi02@gmail.com
}

\usepackage{bibentry}
\begin{document}

\maketitle

\begin{abstract} 

Vision-based autonomous urban driving in dense traffic is quite challenging due to the complicated urban environment and the dynamics of the driving behaviors. Widely-applied methods either heavily rely on hand-crafted rules or learn from limited human experience, which makes them hard to generalize to rare but critical scenarios. In this paper, we present a novel CAscade Deep REinforcement learning framework, CADRE, to achieve model-free vision-based autonomous urban driving. In CADRE, to derive representative latent features from raw observations, we first offline train a Co-attention Perception Module (CoPM) that leverages the co-attention mechanism to learn the inter-relationships between the visual and control information from a pre-collected driving dataset. Cascaded by the frozen CoPM, we then present an efficient distributed proximal policy optimization framework to online learn the driving policy under the guidance of particularly designed reward functions. We perform a comprehensive empirical study with the CARLA \textit{NoCrash} benchmark as well as specific obstacle avoidance scenarios in autonomous urban driving tasks. The experimental results well justify the effectiveness of CADRE and its superiority over the state-of-the-art by a wide margin. \footnote{\Blue{\url{https://github.com/BIT-MCS/Cadre}}}

\end{abstract}

\section{Introduction}
\label{sec:intro}

Autonomous driving~\cite{dickmanns2002development,leonard2008perception,chen2015deepdriving,bojarski2016end,codevilla2018end} has been widely studied over the last few decades and attracted increasing interest due to its huge potential to change the way people travel, while still remaining many technical barriers that must be conquered. Urban driving is one of the most challenging problems due to the high complexity and dynamics of the urban environment (e.g., surrounding vehicles and moving pedestrians). Given the start and target position, the goal of autonomous urban driving is to successfully complete the route within limited time and meet the requirements of pre-defined conditions, such as no collisions.
Under complex urban environment and traffic conditions, it is hard for classic rule-based control approaches to handle all corner cases and avoid violations during driving.

Recently, significant progress has been made for Imitation Learning (IL)~\cite{pomerleau1991efficient,ng2000algorithms}, which enables an agent to mimic the behaviors of human experts. 
For vision-based autonomous driving, benefiting from IL, quite a few works focus on developing an end-to-end behavior cloning system~\cite{codevilla2018end, sauer2018conditional,codevilla2019exploring}, which directly learns a mapping from the raw observations to the control actions, e.g., steer, brake, and throttle. However, as mentioned in~\citet{codevilla2019exploring}, there are several key limitations in IL for vision-based autonomous driving:

\begin{itemize}
  \item [1)] 
The collected dataset from human experts is biased since most of the data are simple behaviors like following the lane. Consequently, the diversity of the data decreases as the dataset grows, which reduces the generalization ability of the agent.    
  \item [2)]
  IL methods suffer from the \emph{distribution shift}, i.e., the training data from experts and the testing observations in the environment are not independent and identically distributed~\cite{bottou2010large}.
\end{itemize}

Deep Reinforcement Learning (DRL) is another promising technique for model-free control and has achieved remarkable success in various complex tasks such as video games and robot control~\cite{van2016deep,fujimoto2018addressing,haarnoja2018soft}. With the guidance of reward signals, DRL methods can alleviate the distribution shift problem through continuously interacting with the environment and effectively correcting current control policies. Some prior works~\cite{dosovitskiy2017carla, kendall2019learning, sallab2017deep} have presented DRL-based autonomous driving frameworks, which, however, can only achieve high performance in simple scenarios, like steering control in an empty town, but fail in dense traffic scenarios. This may be due to two aspects: 
1) The urban environment is very complex, especially in dense traffic containing many vehicles and pedestrians, which requires stronger perception and generalization ability of the DRL agent; 
2) Unlike other control tasks (e.g., video games), the DRL agent is hard to evaluate its derived actions during the long-term and complicated behaviors only through a simple sparse reward signal. 

In this work, we present a novel CAscade Deep REinforcement learning framework, CADRE, to realize model-free vision-based autonomous urban driving. 
CADRE consists of two cascade learning stages: 
1) By leveraging the co-attention mechanism~\cite{lu2016hierarchical} to model the inter-relationships between visual and control information, an offline multi-task learning stage aims to train a Co-attention Perception Module (CoPM) to derive representative latent features from current raw observations.
2) With the help of the frozen CoPM and a carefully designed reward function, an online policy learning stage aims to learn optimal driving policies using distributed Proximal Policy Optimization (PPO)~\cite{schulman2017proximal}.
To the best of our knowledge, we are the first to present a successful online DRL agent on urban driving, particularly with dense traffic handling. 
We summarize our main contributions in the following:
1) We present a novel cascade deep reinforcement learning framework, enabling vision-based autonomous driving in complicated urban environments. 
2) We propose a co-attention perception module (CoPM), which captures the inter-relationships between visual and control information, to better support the environment perception. 
3) A sequential model is leveraged to capture the temporal correlations among the frame sequence, and further contributes to the decision-making for long-term driving.
4) The effectiveness and superiority of our framework are well justified by extensive experiments on \textit{CARLA NoCrash} benchmark and specific obstacle avoidance scenarios in the CARLA simulator~\cite{dosovitskiy2017carla}.

\section{Related Work}
\label{sec:related}

\textbf{Vision-based Urban Autonomous Driving.}
Vision-based urban driving requires an autonomous agent to finish the routes successfully mainly based on visual information. 
The earliest vision-based autonomous vehicles date back to \citet{dickmanns1987autonomous, pomerleau1989alvinn}. 
As one of the most commonly used method to train an autonomous agent, Imitation learning (IL) has been applied in many previous works~\cite{pomerleau1989alvinn, bojarski2016end,codevilla2018end,sauer2018conditional,codevilla2019exploring,chen2020learning, kim2020multi} by mimicking expert demonstrations.
ALVINN~\cite{pomerleau1989alvinn} and DAVE-2~\cite{bojarski2016end} learned a lane following policy which can not take a specific turn and need the driver to take over.
More recently, CIL~\cite{codevilla2018end} trained an imitation learning agent conditioned on high-level commands, making the vehicle  respond to navigational commands.
The following work CILRS~\cite{codevilla2019exploring} proposed the \textit{NoCrash} benchmark discussing the key limitations and further improved CIL by using a deeper residual architecture and a speed regularization.
ChauffeurNet~\cite{bansal2018chauffeurnet} leveraged synthesized data and additional imitation losses for dense urban driving and complex scenarios.
DA-RB+~\cite{prakash2020exploring} proposed an on-policy data aggregation and sampling techniques in the context of dense urban driving.
Recently, ~\citet{zhang2021end} trained an IL agent with the supervisions from an RL coach and BEV image ground-truths.
In this work, we use behavior cloning tasks to help the representative feature extraction from raw observations for subsequent DRL-based agent rather than controlling the vehicle directly.

Deep Reinforcement Learning (DRL) methods try to learn a robust policy via trial and error.
CIRL~\cite{liang2018cirl} fine-tuned a DDPG~\cite{lillicrap2015continuous} agent based on an imitation agent.
~\citet{kendall2019learning} was the first work that applied a DRL agent on the real car.
Having a similar goal,~\cite{osinski2020simulation} transferred a DRL agent trained in CARLA simulator~\cite{dosovitskiy2017carla} to a real-world vehicle.
There are several recent works~\cite{jaafra2019seeking,chen2019model,toromanoff2020end} close to CADRE.
Specifically,~\citet{chen2019model} adapted and compared state-of-the-art model-free deep RL algorithms DDQN~\cite{van2016deep}, TD3~\cite{fujimoto2018addressing} and SAC~\cite{haarnoja2018soft} to the complex autonomous urban driving.
IARL~\cite{toromanoff2020end} was the first DRL agent that can handle complex urban environment including the traffic light detection at intersections.
After that, DeRL~\cite{huang2021deductive} improved the policy learning through predicting future transitions.
Different from the above methods,
we propose an on-policy DRL method with a different perception module to successfully handle dense urban traffic.

\textbf{Attention Mechanisms.}
Attention mechanisms have been widely studied and achieved tremendous success in various tasks (e.g., classification, recognition and machine translation), as in~\citet{mnih2014recurrent,jaderberg2015spatial,cao2015look,vaswani2017,wang2017residual,wang2018non}.
Trained in an end-to-end way, attention can help the neural networks focus on a subset of inputs or latent features which is relatively important for the tasks.
DANet~\cite{Fu2019} proposed the spatial and channel-wise self-attention to integrate local features and global dependencies for scene segmentation.
Specifically, there is a growing number of methods using co-attention mechanisms in vision-and-language tasks like visual question answering~\cite{lu2016hierarchical,nguyen2018improved,wu2018you}.
Co-attention siamese network (COSNet)~\cite{lu2019see} tried to mine the global correlations and scene context among video frames for unsupervised video object segmentation.
These above methods suggest that co-attention is a good way to capture the relationships between different modalities.
There are several works~\cite{chen2017brain, kim2017interpretable,cultrera2020explaining} using the attention mechanisms in autonomous driving, but they did not use the co-attention between different modalities and mainly focused on the visual attention to improve the interpretability of the models.
Different from the existing methods, we introduce control information from behavior cloning tasks and use co-attention to build the correlations between the visual and control information, which can help the subsequent training of the DRL-based agent.
\section{Methodology}\label{sec_method}

\begin{figure}[!t]
	\centering
	\includegraphics[width=1\columnwidth]{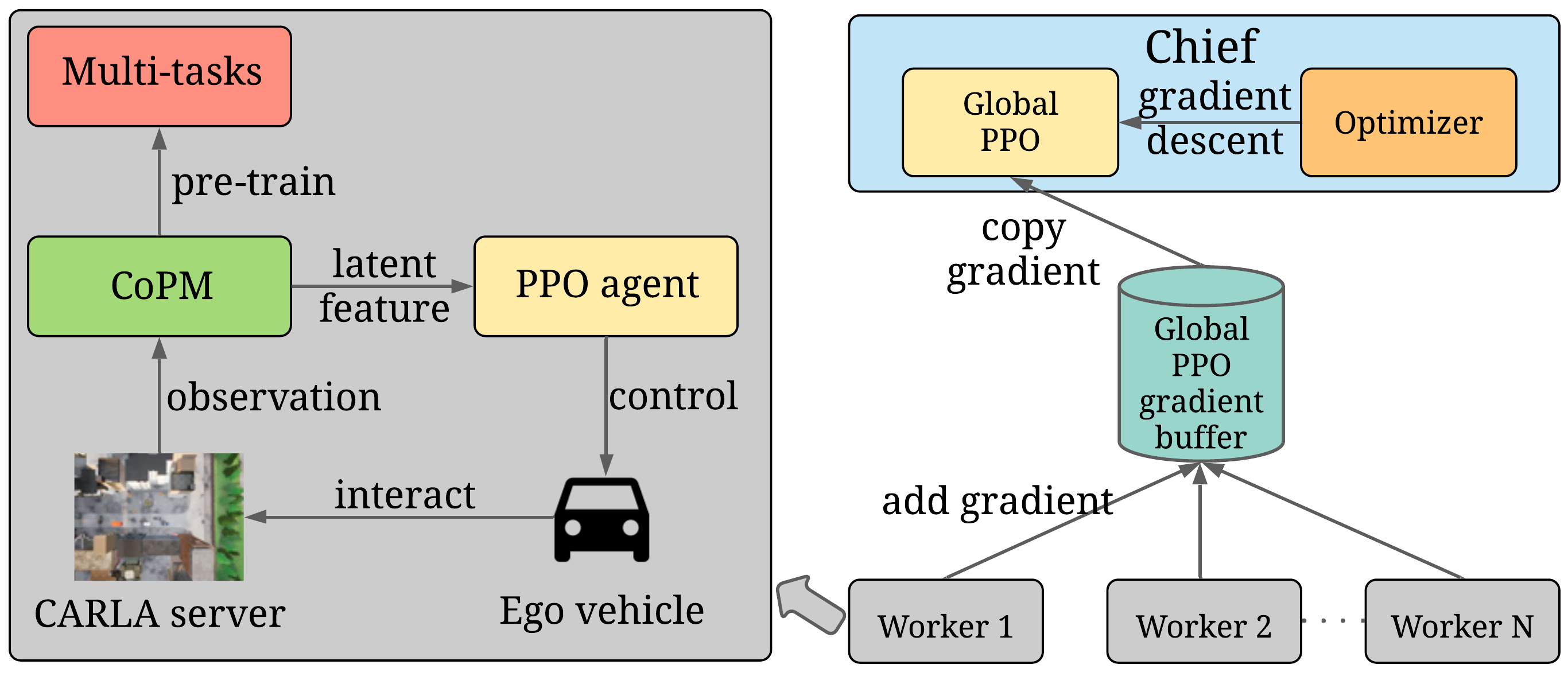}
	\caption{Overview of the CAscade Deep REinforcement learning framework (CADRE).}
    \label{fig_overview}
\end{figure}

In this section, we describe the cascade deep reinforcement learning framework, CADRE, as shown in Figure~\ref{fig_overview}.
Specifically, CADRE consists of two core components: 
1) Co-attention Perception Module (CoPM) leverages the co-attention mechanism to model the inter-relationships between the visual and control information for multi-task learning and provides representative latent features to the PPO agent.
2) Distributed Proximal Policy Optimization (PPO) agent is trained to make decision for the ego vehicle based on the latent features from CoPM using the proposed reward function. 

At timestamp $t$, CADRE receives the observation $\bobs = \langle \bcamera, \bmea, \bway \rangle$ from CARLA~\cite{dosovitskiy2017carla}, where $\bcamera \in \R^{3 \times H \times W}$ is an RGB image of height $H$ and width $W$ from the front-view camera, 
$\bmea$ is a low-dimensional vector referred to as measurements containing information of the ego vehicle (e.g., position, orientation, and speed), 
and $\bway = \{w_1, \cdots, w_K\}$ is a waypoint sequence given by the built-in route planner that the vehicle would follow. 
During the inference, given current observation $\bobs$, the pre-trained Co-attention Perception Module (CoPM) extracts representative latent features $\bz$ for the PPO agent.
Afterward, the PPO agent makes a decision on the steer, throttle, and brake values $\ba = [a^{st}, a^{th}, a^{br}]$ for the ego vehicle. 

\begin{figure}[!t]
	\centering
	\includegraphics[width=1\columnwidth]{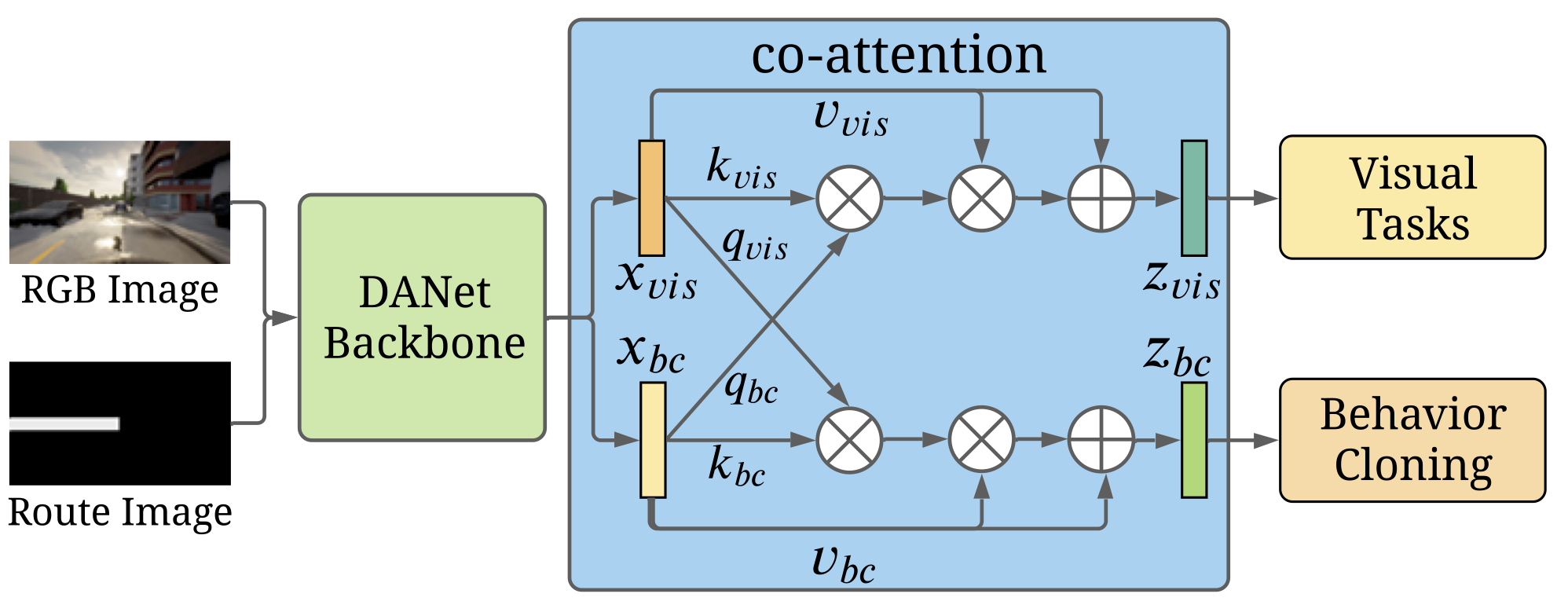}
	\caption{Overview of the Co-attention Perception Module (CoPM). Note that $\otimes$ represents the multiplication operation and $\oplus$ represents the addition operation.}
    \label{fig_CoPM}
\end{figure}

\subsection{Offline Multi-task Learning for Perception}
\label{Section:Environment Perceptron}

\textbf{Observation Preprocessing.}
Given the current observation $\bobs = \langle \bcamera, \bmea, \bway \rangle$, 
one issue is that the camera image $\bcamera$, measurement vector $\bmea$ and waypoint sequence $\bway$ have different structures. 
To fuse the raw observations, 
we propose to preprocess measurements $\bmea$ and waypoint sequence $\bway$ to a route image $\broute$ with the same size as the RGB image $\bcamera$.
Fixing the ego vehicle at the center position of the route image, we calculate the coordinates of each waypoint and connect each pair of two adjacent waypoints by a line.
Afterward, we concatenate the route image $\broute$ and camera image $\bcamera$ in channel and pass it into the CoPM.

\textbf{Co-attention Perception Module.}
As shown in Figure~\ref{fig_CoPM}, the CoPM, trained for the multitask learning~\cite{caruana1997} in an offline supervised way, aims to provide representative features $\bz$ of raw complex observations to the PPO agent.
Taking DANet~\cite{Fu2019} as the backbone, CoPM extracts features containing both the spatial and channel interdependencies from the observations by leveraging the position attention and channel attention.
Here, to gain as much visual and decision-making information as possible, CoPM has a vision branch for the visual tasks as well as a behaviour cloning branch for the control tasks, which differs from~\citet{toromanoff2020end} predicting implicit affordance~\cite{sauer2018conditional}. 
However, in most scenarios of autonomous driving, it is not necessary to focus on all global contexts of the observations (e.g., what is in the middle of the road is usually more important than the scenes on both sides). 
Thus, we should pay more attention to the information that influence the vehicle decision-making.
Towards handling the above limitation, it is crucial to capture the inter-relationships between the visual and control information and thus we propose the co-attention bridging the vision and behaviour cloning branches.

Given the input features $\bx_{vis}$ and $\bx_{bc}$ from DANet backbone for vision and behaviour cloning branches, we obtain the keys, queries and values $\bk_{vis}, \bk_{bc}, \bq_{vis}, \bq_{bc}, \bv_{vis}, \bv_{bc} \in R^{d_{att} \times 1}$ respectively through non-linear blocks by
\begin{align}
\label{Equation: co_att_kqv}
    \bk & = \sigma(\bW_{\bk} x + \bb_{\bk}), \\
    \bq & = \sigma(\bW_{\bq} x + \bb_{\bq}), \\
    \bv & = \sigma(\bW_{\bv} x + \bb_{\bv}),
\end{align}
where $\bk \in \{\bk_{vis}, \bk_{bc}\}$, $\bq \in \{\bq_{vis}, \bq_{bc}\}$, $\bv \in \{\bv_{vis}, \bv_{bc}\}$, $\sigma$ is the activation function, and $\bW, \bb$ are learnable parameters of the non-linear blocks.
Then we calculate the scaled dot-product attention~\cite{vaswani2017} for both branches by
\begin{align}
\label{Equation: co-att_vis}
    \ba_{vis} = \frac{\bq_{bc} \bk_{vis}^T}{\sqrt{d_{att}}}, & \ \ \ba_{bc} = \frac{\bq_{vis} \bk_{bc}^T}{\sqrt{d_{att}}}, \\
    \hat{\ba}_{vis} = \mbox{softmax}(\ba_{vis}), & \ \ \hat{\ba}_{bc} = \mbox{softmax}(\ba_{bc}).
\end{align}

Different from self-attention~\cite{vaswani2017}, we multiply $\bk_{vis}$ by $\bq_{bc}$ and $\bk_{bc}$ by $\bq_{vis}$.
For vision branch, the co-attention is calculated from $\bq_{bc}$ and $\bk_{vis}$, which aims to focus on the visual information important for the decision making. 
For behaviour cloning branch, we use the symmetric attention queried by $\bq_{vis}$, which aims to focus on decision-making information important for the visual tasks.
To keep the original information for both branches, we add the values $\bv_{vis}, \bv_{bc}$ to the attention-weighted sum and then concatenate them to latent feature $\bz$ for the PPO agent in the second stage.
\begin{align}
\label{Equation: co-att_z}
    \bz_{vis} & = \bv_{vis} + \sum_{i=1}^{d_{\bz}}\hat{\ba}_{vis,i} \bv_{vis,i}, \\
    \bz_{bc} & = \bv_{bc} + \sum_{i=1}^{d_{\bz}}\hat{\ba}_{bc,i} \bv_{bc,i}, \\
    \bz & = [\bz_{vis}, \bz_{bc}].
\end{align}

The vision branch consists of three heads $H_{\rrou}$, $H_{\rcam}$ and $H_{\rlig}$ for route image reconstruction, semantic segmentation for camera image and light states classification respectively, to mine the semantic context and extract the visual information in observation $\bobs$.
We use mean square error (MSE) loss for the reconstruction, and cross-entropy loss for the semantic segmentation and classification tasks:
\begin{align}
\label{Equation: L_visual}
    \mL_{\rrou} & = {\left\| \by_{\rrou} - H_{\rrou}(\bz_{vis}) \right\|}_2, \\
    \mL_{\rcam} & = - \sum_{c=1}^{C_{\rcam}} \by_{\rcam} log H_{\rcam}(\bz_{vis})), \\
    \mL_{\rlig} & = - \sum_{c=1}^{C_{\rlig}} \by_{\rlig} log H_{\rlig}(\bz_{vis})),
\end{align}
where $\by_{\rrou}, \by_{\rcam}$ and $\by_{\rlig}$ are the ground-truths, $C_{\rcam}$ and $C_{\rlig}$ are the number of categories for camera image semantic segmentation and light state classification.

The behaviour cloning branch consists of two heads $H_{\rste}$ and $H_{\rthro}$ for the regression of the steer and throttle. 
We use MSE loss for both regression tasks:
\begin{align}
\label{Equation: L_bc}
    \mL_{\rste} & = {\left\| \by_{\rste} - H_{\rste}(\bz_{bc}) \right\|}_2, \\
    \mL_{\rthro} & = {\left\| \by_{\rthro} - H_{\rthro}(\bz_{bc}) \right\|}_2,
\end{align}
where $\by_{\rste}, \by_{\rthro}$ are the ground-truths for steer and throttle.
Combining all the above terms together, we train the CoPM by performing multi-task learning and minimizing the full objective $\mL_{full}$ below:
\begin{align}
\label{Equation: L_full}
    \mL_{\rfull} = & \lambda_{\rrou} \mL_{\rrou} + \lambda_{\rcam} \mL_{\rcam} + \lambda_{\rlig} \mL_{\rlig} \\ 
    & \nonumber + \lambda_{\rste} \mL_{\rste} + \lambda_{\rthro} \mL_{\rthro},
\end{align}
where $\lambda_{\rrou}, \lambda_{\rcam}, \lambda_{\rlig}, \lambda_{\rste}, \lambda_{\rthro}$ are the weights for each loss and are set to 0.5, 1.0, 0.1, 0.1, 0.1, respectively, in our implementation.

\textbf{Data Diversity Augmentation.}
To train the CoPM, we pre-collect a dataset in the CARLA simulator.
But we find that the CoPM pre-trained on the dataset collected using the built-in autopilot can not generalize well on observations that the PPO agent meets during the second stage.
Because the car controlled by autopilot can always perform lane following and turning perfectly and know when to brake to avoid a collision, while the PPO agent usually deviates from the route or hits objects before it converges to the optimal driving policy.
In order to collect imperfect samples and enrich the diversity of samples, we propose to add random noise to the decision made by autopilot with a probability of 0.7: 
\begin{align}
\label{Equation: nosiy_steer}
    a^{st} & := a^{st} + \lambda_{\rst} (2 \cdot noise - 1), \\
    a^{th} & = 0.75, \ \text{if} \ a^{th} < 0.3,
\end{align}
where $a^{st}$ and $a^{th}$ are the steer and throttle returned by the autopilot,  $\lambda_{\rst}$ is set to 10 in our implementation and $noise$ is a variable from uniform distribution $U(0,1)$.
Note that $\ba^{st}$ will be clipped to [-1,1] once it is out of range.
\subsection{Distributed Proximal Policy Optimization for Control}
Given the latent features $\bz$ at time timestamp $t$ from the frozen CoPM, to emphasize the measurement vector $\bmea$ containing the steer, throttle and brake values at timestamp $t-1$, ego vehicle's linear velocity $v$, route deviation degree $\theta$ and route deviation distance $d$ at timestamp $t$, which are important for making decision, we explicitly concatenate $\bmea$ and $\bz$ to $\hat{\bz}$ as the state for the PPO agent.
To help the agent learn a good policy (e.g, decelerate quickly to avoid collision), given the continuous action space $\ba = [a^{st}, a^{th}, a^{br}]$ for the steer, throttle, and brake, we discretize them resepectively. In CARLA, a vehicle's basic movement is controlled by three different values, i.e., steer value $a^{st} \in [-1.0, 1.0]$, throttle value $a^{th} \in [0.0, 1.0]$ and brake value $a^{br} \in [0.0, 1.0]$. Throttle value will be automatically set to 0 if brake value is larger than 0. 
For the steer space, we discretize it into 33 bins from -1 to 1 evenly, where -1 corresponds to fully turning left and 1 corresponds to fully turning right. 
For the throttle and brake space, we merge them together and discretize it into 3 different actions:
\textit{accelerate} (throttle is 0.6 and brake is 0),
\textit{move forward} (throttle is 0 and brake is 0), and 
\textit{decelerate}  (throttle is 0 and brake is 1).

\textbf{Temporal Dependencies.}
Through the frozen CoPM, the latent feature $\hat{\bz}$ contains the inter-relationships between the visual and behavior cloning information and current measurements from multi-sensors on the ego vehicle.
However, we have only fully exploited the information in the current observation, which is not enough since driving is a long-term behavior.
To capture the temporal dependencies among consecutive frames, we leverage a commonly used sequential model LSTM~\cite{hochreiter1997long} to integrate information from eight consecutive frames:
\begin{align}
\label{Equation: lstm}
    \hat{\bz_t} := LSTM(\hat{\bz}_{t-7:t}),
\end{align}
where the subscript $t$ represents the timestamp.

\textbf{Reward Shaping.}
Our reward formulation includes two parts: 1) The agent obtains a \emph{sparse reward} when a pre-defined event is triggered;
2) The agent obtains a \emph{dense reward} at every timestamp.
For the sparse reward, we define four types of bad events giving a penalty reward to the agent: 
1) collision static objects,
2) collision vehicles or pedestrians,
3) vehicle blocked, and
4) route deviation.
We also define a good event giving a bonus reward: complete the route successfully.
Note that once an event happens, the current episode ends immediately and a new episode begins.
Although the sparse reward is critical, it can not help the model evaluate the local action due to its sparsity.
Therefore, to address the sparse reward problem, we define the dense rewards consisting of a deviation degree reward $r^{\theta}$, a deviation distance reward $r^{d}$ and a velocity reward $r^v$. 
The deviation degree reward $r^{\theta}$ is calculated as follows:
\begin{equation}
  r^{\theta}=\max(0, 1-\frac{\theta}{\theta_{max}}),
\end{equation} 
where $\theta$ is the route deviation degree and $\theta_{max}$ is the maximum threshold set to 90$^\circ$.
It indicates that $r^{\theta}$ decreases with the increasing of $\theta$. 
If $\theta$ is larger than the maximum allowed value, then the agent will get a minimum reward of 0.
The deviation distance reward $r^d$ is calculated as follows:
\begin{equation}
  r^{d}=\max(0, 1-\frac{d}{d_{max}}),
\end{equation}
where $d$ is the route deviation distance and $d_{max}$ is the maximum threshold set to 2.5m.
It indicates that $r^{d}$ decreases with the increasing of $d$.
If $d$ is larger than the maximum allowed value, then the agent will get a minimum reward of 0.
The velocity reward $r^v$ is calculated as follows:
\begin{equation}
r^v=\left \{
\begin{aligned}
\min(1, \frac{v}{v_{min}}), & & \textrm{if $v < v_{target}$,}\\
\max(0, 1-\frac{v-v_{target}}{v_{max}-v_{target}}), & & \textrm{otherwise.}
\end{aligned}
\right.
\end{equation}
The velocity reward indicates that the ego vehicle's speed should follow a target speed range to achieve an optimal speed control. When there are no dynamic objects (i.e., vehicles and pedestrians) detected, we set $v_{min}, v_{max}$ as the minimum and the maximum suggested velocity according to the traffic rules respectively, and $v_{target}$ as the target velocity defined by $v_{target}=(v_{min}+v_{max})/2$. When an obstacle exists ahead, we set $v_{min}$ as $v$, $v_{target}$ as the distance value to the obstacle, $v_{max}$ as the maximum detection distance of the obstacle sensor. This setting indicates that the ego vehicle should shift down when it approaches the obstacle.

\textbf{Distributed Training.}
Following~\citet{schulman2017proximal}, the clipped surrogate objective of the PPO agent is
\begin{gather}
    r_t =\frac{\pi_(\ba_t|\hat{\bz}_t)}{\pi_{old}(\ba_t|\hat{\bz}_t)}, \\
    \mL_{\rclip} = \mathbb{E}_t[\min(r_t A_t, clip(r_t,1-\epsilon, 1+\epsilon) A_t)],\label{clip objective}
\end{gather}
where $\hat{\bz}_t$ is the latent features from the LSTM, $\ba_t$ is the predicted action, $A_t$ is the advantage function (i.e., the accumulated discounted rewards minus the state value $V(\hat{\bz}_t)$ in PPO), and $r_t$ is the probability ratio with respect to old policy $\pi_{old}$ and updated policy $\pi$. 
The $clip$ operation removes the incentive for moving $r_t$ out of interval [$1-\epsilon$, $1+\epsilon$], which makes the training process more stable. The goal of the second stage for CADRE is to minimize $\mL_{clip}$ in order to derive an optimal driving policy $\pi$.
As shown in Figure~\ref{fig_overview}, we use a chief-worker distributed architecture to achieve distributed training. 
Each worker contains a model and a CARLA server at local, where the PPO agent collects driving experience and computes the gradients independently.
Once the global PPO gradient buffer gathers all the local gradients from each worker, the global PPO in the chief process performs a gradient descent and copy the model parameter to each local worker.
Afterward, a new training iteration begins.
\section{Experiments}
\label{sec:exp}

We conducted all the experiments on the latest CARLA simulator~\cite{dosovitskiy2017carla} version 0.9.10. Through the detailed experiments, we aim to answer the following questions:
1) How does CADRE perform in urban driving compared with other baselines, especially on dense traffic condition? 
2) For the perception subtask, is it beneficial to the training of the DRL-based agent by introducing the control information and the corresponding inter-relationships?
3) For the control subtask, what are the strengths and weaknesses of on-policy DRL methods?
What insight does CADRE provide to alleviate the weaknesses?

\subsection{Experimental Setup}
\textbf{Simulator Environment.} 
CARLA~\cite{dosovitskiy2017carla} contains high-fidelity maps with many professionally designed static objects (e.g., buildings and traffic signs), as well as dynamic objects (e.g., dozens of types of vehicles and pedestrians with different appearances).
Overall it provides a convenient and realistic environment for experiments in autonomous driving.

\textbf{Evaluation Metrics.} 
In all of the following experiments, we use two metrics referred as \emph{average route completion ratio} and \emph{success rate}~\cite{codevilla2019exploring} to evaluate the training process and testing results respectively. 
Route completion is the ratio of the successful distance travelled by the vehicle to the total length of the route.
The \emph{average route completion ratio} is computed by averaging the route completion ratio results of all training routes in the training process. 
As defined in~\citet{codevilla2019exploring}, the success rate is $1$ for each route if the ego vehicle completes the required route in a limited time range with no collisions, otherwise is $0$.

\textbf{\textit{NoCrash} Benchmark.}
\textit{NoCrash} benchmark is proposed in~\citet{codevilla2019exploring} to evaluate the autonomous driving policy under various urban conditions. Basically, it contains three different traffic conditions with increasing difficulty levels as 
\textit{empty} (no dynamic objects),
\textit{regular} (medium number of pedestrians and vehicles), 
and \textit{dense} (a large number of pedestrians and vehicles).
Besides, it defines 6 kinds of weathers and 25 routes in Town01 for training and Town02 for testing. We train our autonomous agent in the training town and evaluate the performance under the testing weathers and testing town respectively.

\textbf{Obstacle Avoidance Scenarios for Evaluation.} \textit{NoCrash} benchmark evaluates the agent on a long route, during which every scenario is formed with the random behaviors of the vehicles and pedestrians, such as pedestrians crossing the road, which is unreproducible in some sense. Furthermore, there are 27 kinds of vehicles and 26 kinds of pedestrians in CARLA 0.9.10. The differences in the appearance of different objects (e.g., ordinary cars and large trucks) can cause the agent to act differently, which is not taken into account in \textit{NoCrash} benchmark.

Therefore, to reduce the randomness and to evaluate the obstacle avoidance performance and inertia problem \cite{codevilla2019exploring} more accurately, we designed a set of short routes with obstacle avoidance scenarios with each kind of vehicle and pedestrian in Town01 and Town02 (i.e, there are 27 and 26 kinds of routes for vehicle and pedestrian avoidance).
Note that all the routes are only for evaluation.
Same as in \textit{NoCrash} benchmark, we calculate the success rates across all routes.

\emph{Vehicle avoidance scenarios.} 
When the ego vehicle arrives at the trigger point, a stationary tool vehicle will be generated 20 meters ahead for a period of time in order to block the ego vehicle.
The ego vehicle has to stop in time and continue driving after the tool vehicle disappears.

\emph{Pedestrian avoidance scenarios.} 
When the ego vehicle arrives at the trigger point, a pedestrian will be generated 20 meters ahead at the sidewalk in order to cross the road and block the ego vehicle.
The ego vehicle has to stop in time and continue driving after the pedestrian crosses the road.

\textbf{Dataset for CoPM.}
In the first stage, for training the CoPM, we collected a large and diverse dataset following the 25 training routes under three conditions defined in \textit{NoCrash} benchmark from CARLA, using the built-in autopilot with additional random noise as mentioned before.
The dataset has 315,545 samples, each containing a $256 \times 144$ monocular RGB image, a waypoint sequence, vehicle measurements and the ground truths for the visual and behavior cloning tasks. 

\textbf{Route and Scenario Augmentation in Training.} The predefined routes and traffic scenarios in \textit{NoCrash} is not sufficient for DRL policy training. Therefore, we use a route and scenario augmentation method to boost a better driving policy. First, we cut the 25 long training routes into 112 short routes where there are even number of routes under each command. After that, we predefine the vehicle and pedestrian avoidance scenarios along these short routes in order to obtain a good driving policy. Last, we resume the ego vehicle from the failure location in each location rather than choosing a random start point to increase the number of samples which are meaningful for policy learning. 
It takes totally 3.2M training samples for CADRE to converge at a stable performance, which is enormously fewer than the 20M training samples required by the off policy DRL method IARL~\cite{toromanoff2020end}. 
Apparently, it is an important strength of on policy method that it needs less training samples for converge.

\subsection{Results on \textit{NoCrash} Benchmark}
We compared CADRE with the state-of-the-art model-free reinforcement learning method IARL~\cite{toromanoff2020end} and imitation learning method LBC~\cite{chen2020learning} on CARLA 0.9.10. Results are taken from~\citet{chen2021learning}. Note that on CARLA 0.9.10, IARL released a strong model trained on all towns and all weathers. In that case, this model does not have held-out testing weathers and thus we can only get the results under the training weathers.

Table~\ref{Tab:nocrash} shows the quantitative results in terms of the success rate on \textit{NoCrash} benchmark.
For answering the first question, we observe that, on version 0.9.10, CADRE consistently outperformed LBC under test weathers on all traffic conditions. 
Especially, CADRE achieved 82 and 61 success rates in training and testing town respectively under dense traffic and training weathers, with 19 and 28 success rates higher than the performance achieved by IARL.
It demonstrates that CADRE can drive the vehicle carefully and keep the high performance on the dense traffic condition. 
Even driving on the testing unseen weather, CADRE can keep the high success rates on all conditions via a good policy, while LBC dropped its performance dramatically. It is worth noting that the behavior cloning branch (3rd column in Table~\ref{Tab:nocrash}) in the CoPM can also achieve comparable performance on all conditions.
We also provide more detailed reference results for performance comparison between CADRE and the state-of-the-art on different versions of CARLA in the supplemental materials.

\begin{table}[t]
\begin{center}
\resizebox{1.0\columnwidth}{!}{
\begin{tabular}{lll|cccc}
\toprule
Task& Town & Weather & IARL & LBC & CoPM & Ours \\
\midrule
Empty&~& ~ & $85$ & $89$ & $62$ & $\bm{95}$\\ 
Regular&Train & Train&$86$&$87$ & $63$ & $\bm{92}$ \\
Dense&~&~ &$63$ & $75$ & $70$ &$\bm{82}$ \\
\midrule
Empty&~ &~ & $-$ &$60$ & $62$ &$\bm{94}$\\ 
Regular&Train&Test &$-$ & $60$ & $66$ & $\bm{86}$\\
Dense&~& ~&$-$& $54$ & $72$ & $\bm{76}$ \\
\midrule
Empty&~ &~ & $77$ &$86$ & $86$ &$\bm{92}$\\ 
Regular&Test&Train &$66$ & $\bm{79}$ & $70$ & $78$\\
Dense&~& ~&$33$& $53$ & $39$ & $\bm{61}$ \\
\midrule
Empty&~ &~ & $-$ &$36$ & $44$ &$\bm{78}$\\ 
Regular&Test&Test &$-$ & $36$ & $44$ & $\bm{72}$\\
Dense&~& ~&$-$& $12$ & $30$ & $\bm{52}$ \\
\bottomrule
\end{tabular}}
\end{center}
\caption{Results on \textit{NoCrash} benchmark.}
\label{Tab:nocrash}
\end{table}

\subsection{Results on Obstacle Avoidance Scenarios}

Table~\ref{Tab:scenarios} shows the quantitative results in terms of the success rate on our designed routes with specified obstacle avoidance scenarios.
Same as on the \textit{NoCrash} benchmark, all the experiments are built on CARLA 0.9.10.
We repeated the evaluation 3 times for each scenario (i.e., 81 times for vehicle avoidance and 78 times for pedestrian avoidance in total).
In both vehicle and pedestrian avoidance scenarios, CADRE consistently outperformed IARL and LBC, which indicated that CADRE has a better perception ability for the obstacles.

\begin{table}[t]
\begin{center}
\begin{tabular}{l|cc|cc}
\toprule
&  \multicolumn{1}{c|}{Vehicle avoidance} & \multicolumn{1}{c}{Pedestrian avoidance} \\
\midrule
LBC~&  \multicolumn{1}{c|}{$55/81$} & \multicolumn{1}{c}{$73/78$} \\
IARL~ &  \multicolumn{1}{c|}{$69/81$} & \multicolumn{1}{c}{$57/78$} \\
Ours & \multicolumn{1}{c|}{$81/81$} &  \multicolumn{1}{c}{$76/78$} \\

\bottomrule
\end{tabular}
\end{center}
\caption{Results on obstacle avoidance scenarios.}
\label{Tab:scenarios}
\end{table}

\begin{table}[t]
\begin{center}
\resizebox{1.0\columnwidth}{!}{
\begin{tabular}{l|cc|cc}
\toprule
&  \multicolumn{1}{c|}{Vehicle avoidance} & \multicolumn{1}{c}{Pedestrian avoidance} \\
\midrule
Visual-only CoPM & \multicolumn{1}{c|}{$75/81$} &  \multicolumn{1}{c}{$39/78$} \\
CoPM w/o att &  \multicolumn{1}{c|}{$76/81$} & \multicolumn{1}{c}{$68/78$} \\
CoPM &  \multicolumn{1}{c|}{$81/81$} & \multicolumn{1}{c}{$76/78$} \\

\bottomrule
\end{tabular}
}
\end{center}
\caption{Ablation study on the CoPM on obstacle avoidance scenarios.}
\label{Tab:CoPM-scenarios}
\end{table}

\begin{table}[t]
\begin{center}
\resizebox{1.0\columnwidth}{!}{
\begin{tabular}{l|cc|cc}
\toprule
&  \multicolumn{1}{c|}{Vehicle avoidance} & \multicolumn{1}{c}{Pedestrian avoidance} \\
\midrule
Basic CADRE & \multicolumn{1}{c|}{$51/81$} &  \multicolumn{1}{c}{$51/78$} \\
Basic CADRE w/ LSTM	 &  \multicolumn{1}{c|}{$71/81$} & \multicolumn{1}{c}{$61/78$} \\
Complete CADRE &  \multicolumn{1}{c|}{$81/81$} & \multicolumn{1}{c}{$76/78$} \\

\bottomrule
\end{tabular}
}
\end{center}
\caption{Ablation study on the PPO on obstacle avoidance scenarios.}
\label{Tab:PPO-scenarios}
\end{table}

\subsection{Ablation Study}

\begin{figure*}[htp]
    \centering
    \begin{subfigure}[t]{0.9\columnwidth}
        \includegraphics[width=1\columnwidth]{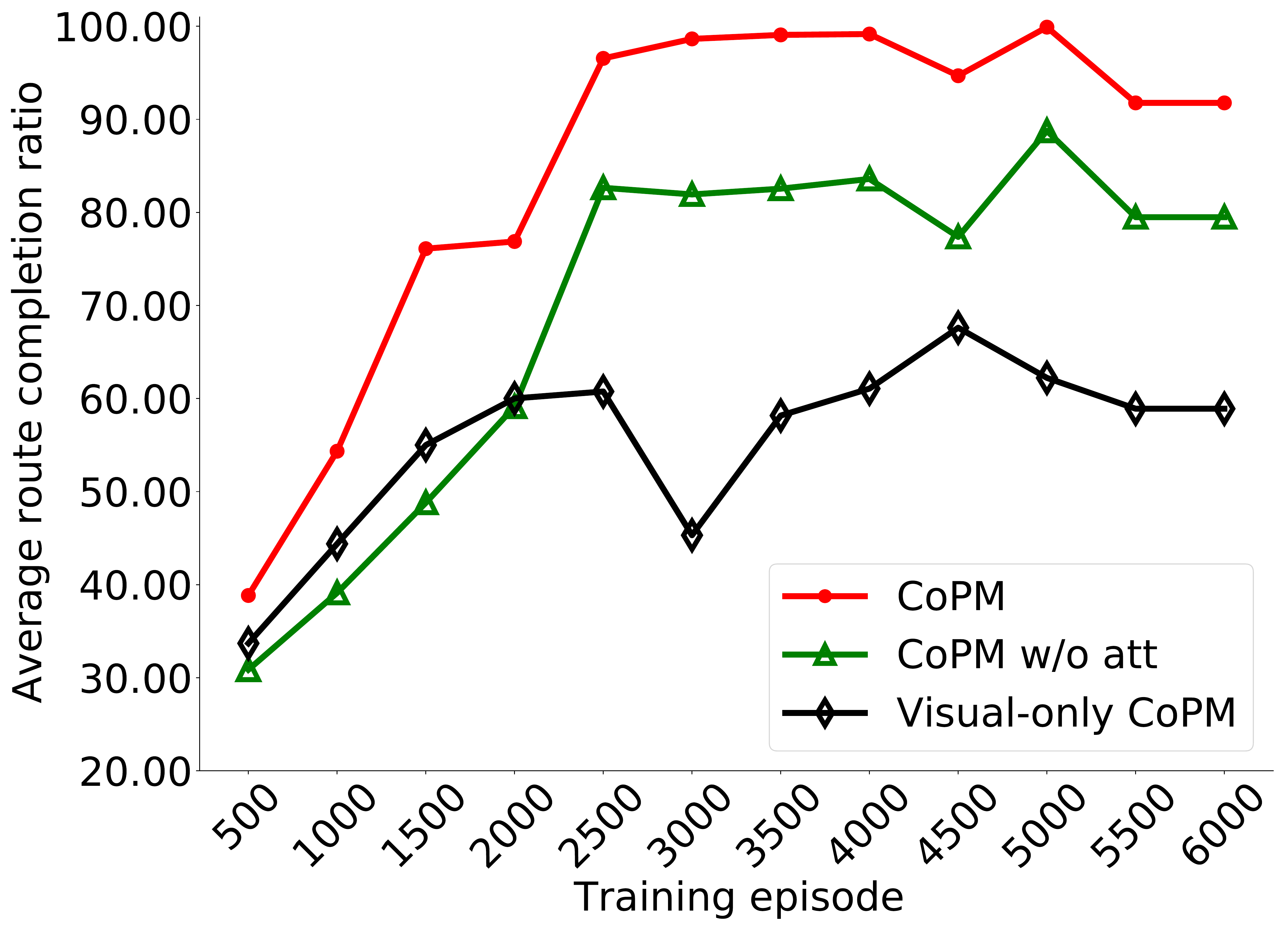}
        \caption{Training curves of the ablation study on the CoPM.}
        \label{fig_CoPM_ablation}
    \end{subfigure}%
    \hspace{3ex}
    \begin{subfigure}[t]{0.9\columnwidth}
        \includegraphics[width=1\columnwidth]{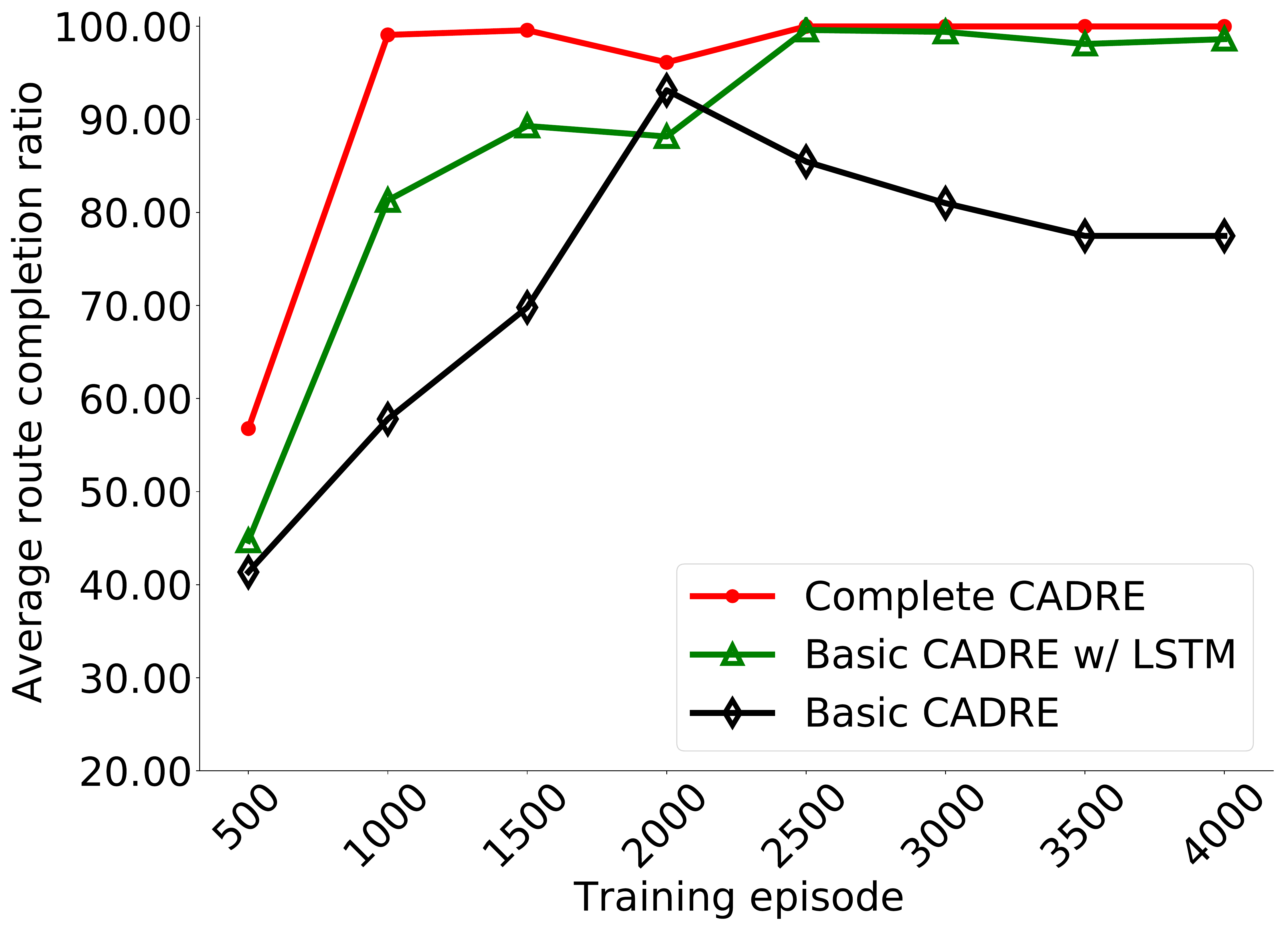}
        \caption{Training curves of the ablation study on the PPO agent.}
        \label{fig_ppo_ablation}
    \end{subfigure}%
\caption{Ablation study for the CADRE.}
\end{figure*}

To better understand the merits of CADRE, we conducted detailed ablation studies on the CoPM and the PPO agent in Figures~\ref{fig_CoPM_ablation} and \ref{fig_ppo_ablation} 
respectively.

\textbf{Ablation Study on the CoPM.}
To answer the second question and verify the effectiveness of the co-attention mechanism and the control information,
we compared the complete CoPM to two simpler baselines: 1) CoPM without co-attention (CoPM w/o att) and 2) Visual-only CoPM that is trained only for visual tasks thus without co-attention.
Figure~\ref{fig_CoPM_ablation} shows the quantitative results on the training process of the PPO agent in terms of the average route completion ratio.
We can observe that the CoPM (red line) achieved a higher average route completion ratio than the CoPM w/o att (green line) and the visual-only CoPM (black line), 
which indicated that CoPM can build the inter-relationships between the visual and control information, and extract more representative latent features to help the learning of the PPO agent. In addition, Table~\ref{Tab:CoPM-scenarios} shows the quantitative results about the CoPM in terms of the success rate on obstacle avoidance scenarios.
The CoPM without co-attention outperformed the visual-only CoPM by relatively large margins, showing the control information from the behavior cloning task is beneficial to the learning of the PPO agent.

\textbf{Ablation Study on the PPO Agent.}
To answer the third question, We built an ablation study to illustrate the contribution of each component in the DRL policy model and demonstrate the training process with \emph{average route completion ratio} metric. We use three settings of CADRE as follows:
1) Basic CADRE: consisting of the CoPM with a vanilla PPO policy model;
2) Basic CADRE w/ LSTM: adding the LSTM module to capture the temporal correlations among frame sequence;
3) Complete CADRE: the basic CADRE w/ LSTM in a distributed training manner with 4 workers.
As shown in Figure~\ref{fig_ppo_ablation}, compared with the basic CADRE, CADRE w/ LSTM converged faster and obtained a better result in \emph{average route completion ratio}, which indicates that the temporal correlations among the consecutive frames are beneficial to the learning.
Clearly, complete CADRE converged faster than all of the baselines and eventually stabilized to an \emph{average route completion ratio} nearly to $100\%$. 

It should be noticed that there is a performance drop for all these methods in episode 1500 -- 2000. Complete CADRE quickly recovers from the performance drop while the basic CADRE remains at a \emph{average route completion ratio} at 78\%. This result indicates that it is possible for an on policy method to crash into a local optimal performance and never recover. However, with the distributed training architecture and our carefully designed reward function, it can recover from the local optimal performance and remain stable.

Table~\ref{Tab:PPO-scenarios} shows the quantitative results about the PPO agent in terms of the success rate on obstacle avoidance scenarios.
We can clearly observe that both the temporal correlations from LSTM and distributed training contribute a lot to the performance improvement on obstacle avoidance scenarios.

\section{Conclusion}
\label{sec:clu}

In this paper, we proposed a novel CAscade Deep REinforcement learning (DRL) framework, CADRE for vision-based autonomous urban driving.
To reduce the complexity of the driving task, CADRE splits it into perception and control subtasks.
In the first stage, we pre-train a Co-attention Perception Module (CoPM), which leverages the co-attention mechanism to build the inter-relationships between the visual and control information, to provide representative latent features from the raw observations to the subsequent agent.
In the second stage, we use an on-policy algorithm, Proximal Policy Optimization (PPO), to train the agent in a distributed way. 
With the careful reward shaping and sequential model LSTM, the PPO agent learns a good policy, which can achieve high success rates in complex urban environment even on the dense traffic condition.
Results on \textit{NoCrash} benchmark and specific obstacle avoidance scenarios demonstrated the effectiveness of CADRE. 

\section{Acknowledgements}

This work was done while the first authors, Yinuo Zhao and Kun
Wu, were interns at Midea Group. This work was supported in part by Shanghai Pujiang Program and the National Research and Development Program of China (No. 2019YQ1700).

\bibliography{Cadre}

\end{document}